\documentclass[letterpaper, 10 pt, conference]{ieeeconf}  % Comment this line out if you need a4paper

\IEEEoverridecommandlockouts                              % This command is only needed if 
\usepackage{graphicx}
\usepackage{amsmath} % assumes amsmath package installed
\usepackage{tikz}
\usepackage[english]{babel}
\usepackage[ruled]{algorithm2e}
\usepackage{floatrow}
\usepackage{comment}
\usepackage{hyperref}
\usepackage{amssymb}
\usepackage{multirow}
\usepackage{tcolorbox}  % For boxed content
\usepackage{fancyhdr}   % For controlling header/footer
\usepackage[letterpaper, margin=0.75in]{geometry} % Adjust margins as needed

\DeclareMathOperator*{\argmax}{argmax}

\newfloatcommand{capbtabbox}{table}[][\FBwidth]

\title{\LARGE \bf
Improving robot navigation in crowded environments using intrinsic rewards
}

\author{Diego Martinez-Baselga, Luis Riazuelo and Luis Montano$^{1}$% <-this % stops a space
\thanks{$^{1}$The authors are with the Robotics, Perception and Real Time Group, Aragon Institute of Engineering Research (I3A), Universidad de Zaragoza, 50018 Zaragoza, Spain.
        {\tt\small \{diegomartinez,riazuelo,montano\}@unizar.es}}% <-this % stops a space
}

\newcommand\copyrighttext{%
  \footnotesize \textcopyright This paper has been accepted for publication at IEEE 2023 International Conference on Robotics and Automation (ICRA). Please, when citing the paper, refer to the official manuscript with the following DOI: 10.1109/ICRA48891.2023.10160876.}
\newcommand\copyrightnotice{%
\begin{tikzpicture}[remember picture,overlay]
\node[anchor=south,yshift=10pt] at (current page.south) {\fbox{\parbox{\dimexpr\textwidth-\fboxsep-\fboxrule\relax}{\copyrighttext}}};
\end{tikzpicture}%
}

\begin{document}
\maketitle
\copyrightnotice

\pagestyle{empty}

%%%%%%%%%%%%%%%%%%%%%%%%%%%%%%%%%%%%%%%%%%%%%%%%%%%%%%%%%%%%%%%%%%%%%%%%%%%%%%%%

\begin{abstract}
Autonomous navigation in crowded environments is an open problem with many applications, essential for the coexistence of robots and humans in the smart cities of the future. In recent years, deep reinforcement learning approaches have proven to outperform model-based algorithms. Nevertheless, even though the results provided are promising, the works are not able to take advantage of the capabilities that their models offer. They usually get trapped in local optima in the training process, that prevent them from learning the optimal policy. They are not able to visit and interact with every possible state appropriately, such as with the states near the goal or near the dynamic obstacles. In this work, we propose using intrinsic rewards to balance between exploration and exploitation and explore depending on the uncertainty of the states instead of on the time the agent has been trained, encouraging the agent to get more curious about unknown states. We explain the benefits of the approach and compare it with other exploration algorithms that may be used for crowd navigation. Many simulation experiments are performed modifying several algorithms of the state-of-the-art, showing that the use of intrinsic rewards makes the robot learn faster and reach higher rewards and success rates (fewer collisions) in shorter navigation times, outperforming the state-of-the-art.

\end{abstract}

%%%%%%%%%%%%%%%%%%%%%%%%%%%%%%%%%%%%%%%%%%%%%%%%%%%%%%%%%%%%%%%%%%%%%%%%%%%%%%%%
\section{Introduction}

Robotic autonomous navigation in dynamic environments is a very complex problem. There is a huge variety of approaches that has been proposed to solve it, but none of them is able to succeed in every single scenario, leading to collisions or very long trajectories.

In the traditional navigation stack, a global planner, which typically is a variant of the A$*$ algorithm \cite{hart1968formal} like as D$*$ Lite \cite{koenig2005fast} or LPA$*$ \cite{koenig2004lifelong}, is used to compute the shortest feasible path between the robot and a selected goal. The local planner is in charge of avoiding collisions with dynamic obstacles and other obstacles that were not considered in the map, and compute the motion of the robot according to its kinodynamic constraints. In the Dynamic Window Approach (DWA) \cite{fox1997dynamic}, a local planner that considered current obstacles poses and robot's kinodynamic is proposed. Other works also consider the velocity of the obstacles, such as the Velocity Obstacle (VO) \cite{fiorini1998motion} or the Dynamic Object Velocity Space (DOVS) \cite{lorente2018model}, which is necessary in dynamic environments. The VO was extended for multi-robot reciprocal collision avoidance, with the Reciprocal Velocity Obstacle (RVO) \cite{van2008reciprocal} or the Optimal Reciprocal Collision Avoidance (ORCA) \cite{berg2011reciprocal}, but need the obstacles to collaborate reciprocally.

Navigating autonomously in crowded environments poses a challenge with applications in areas such as airport halls, crowded streets for delivery, restaurants, and warehouses. Recent efforts to solve this problem have utilized deep reinforcement learning (DRL), which has produced encouraging results that outperform previous model-based algorithms. The exploration and exploitation process is crucial for DRL algorithms, where the policy is refined through exploration (adding randomness to decision-making when the policy is unknown) and exploitation (refining the policy based on what has been learned from the environment). However, in this specific problem, very unknown states may be found at any time of the training process, not only at the beginning of the trajectories and the training, and very rewarding maneuvers could be achieved in states that may not look promising, such as near humans, leading to suboptimal policies and local optima in the optimization of network weights.

\begin{figure}
    \centering
    %\begin{tabular}[b]{c}
         \includegraphics[trim={0.5cm 0.5cm 1.75cm 1.75cm},clip,width=0.49\textwidth]{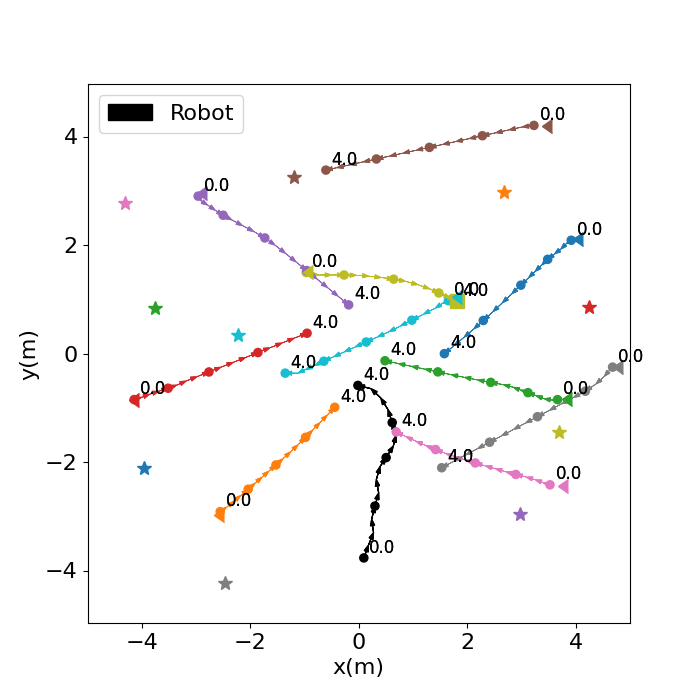}
         %\\
         %\footnotesize{(d) $\epsilon$-greedy best}
     %\end{tabular}
     %\hfill
    %\begin{tabular}[b]{c}
         \includegraphics[trim={0.5cm 0.5cm 1.75cm 1.75cm},clip,width=0.49\textwidth]{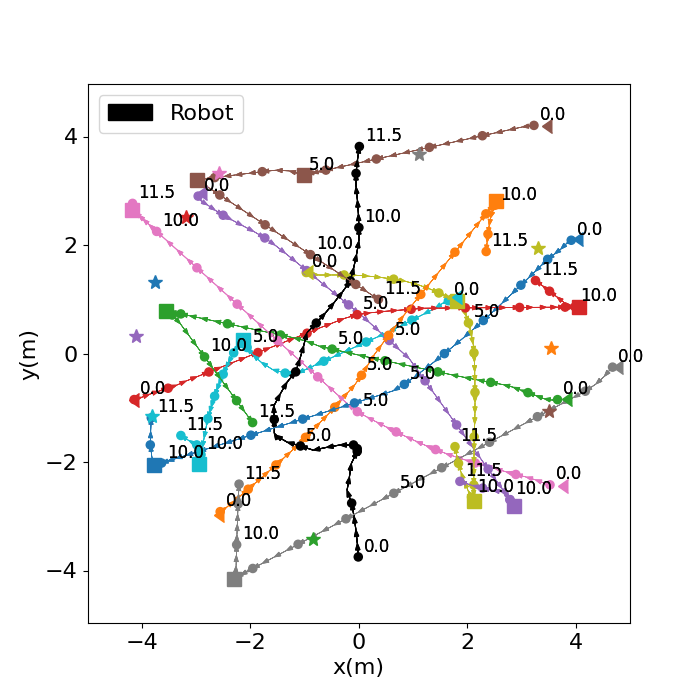} 
         %\\
         %\footnotesize{(h) ICM best}
     %\end{tabular}
        \caption{Comparison shows colliding (left) vs. safe (right) trajectory, obtained by same algorithm without/with intrinsic rewards.}
        \label{fig:motivation-fig}
\end{figure}

In this work, we propose using intrinsic rewards to overcome the exploration problems of the state-of-the-art algorithms and get policies that learn faster and score higher rewards and success rates in shorter times. For example, in Figure~\ref{fig:motivation-fig}, we may see how the robot collides with one of the humans using an algorithm of the state-of-the-art (left), and how it is able to reach the goal in the exact same scenario by using the same algorithm trained with intrinsic rewards (right). To our knowledge, we are the first to introduce smart exploration algorithms in this problem and, specifically, the first to introduce intrinsic rewards. Different approaches are trained and tested with several recent and well-known state-of-the-art algorithms in Section~\ref{sec:experiments}, including results with comparisons between the original and proposed algorithms. The code and videos of the experiments may be accessed at \url{https://github.com/dmartinezbaselga/intrinsic-rewards-navigation.git}.

\section{Background}

\subsection{Crowd navigation}

Robot navigation in crowds is a very challenging problem, due to the fact that it combines the problems of dynamic environments with the complexity added by the interactions between humans, which modify their motion. Simply and reactive approaches fail to navigate in these scenarios, as they do not consider social interactions, modeled in works as Social Force \cite{helbing1995social} and Extended Social Force \cite{jiang2017extended} to predict how the crowd would behave and compute the motion according to it. Moreover, trajectory based approaches are too computationally costly to be applied in real time when the crowd size increase, or are not able to find safe paths \cite{trautman2010unfreezing}.

Recent approaches use DRL to estimate the optimal policy the robot must follow to navigate safely towards the goal. One of the ways they differ among each other is the input (state description) they use. Some approaches use sensor measurements directly to describe the environment \cite{dugas2021navrep} \cite{yokoyama2020autonomous} \cite{shi2019end} \cite{tai2018socially} or abstractions from the measurements \cite{patel2021dwa}. 

There is a big group of works that use the information of the other agents that are in the scenario as the input, including their positions or velocities. In CADRL (Collision Avoidance with Deep Reinforcement Learning) \cite{chen2017decentralized}, an algorithm that outperforms ORCA using DRL in multi-agent collision avoidance was introduced, considering only the closest agent to the robot. Other approaches used Long Short-Term Memory (LSTM) \cite{hochreiter1997long} in order to include more than just one agent (LSTM-RL) \cite{everett2018motion} \cite{everett2021collision}. In Socially Attentive Reinforcement Learning (SARL) \cite{chen2019crowd}, the Human-Robot and Human-Human interaction is modeled and the information is processed using attention models \cite{vaswani2017attention}. Other works predicted attention weights by using graph neural networks, obtaining better performance than the previous motion planners \cite{chen2020relational} \cite{chen2020robot}. In \cite{zhou2022robot}, they propose a method with better results than the ones previously mentioned, tested in the same CrowdNav simulator. It uses the social attention mechanism \cite{vemula2018social} to compute the attention weights and online planning.

\subsection{Exploration in Deep Reinforcement Learning}

In DRL, algorithms try to estimate a policy that maximizes the accumulated discounted reward provided by an environment in an episode. Agents should balance properly between exploration and exploitation to discover the states where the highest reward is obtained and improve the policy.

The most basic and common exploration strategies are $\epsilon$-greedy, which selects the greedy action with a probability of $1-\epsilon$ and a random action otherwise; and Boltzmann (softmax) exploration, which makes the agent choose actions using a Boltzmann distribution over the predicted Q-values. Nevertheless, in complex environments with sparse rewards or where the policy is specially hard to learn (typically real-world environments), those exploration strategies are not enough.

A commonly used approach is including the entropy of the policy to the objective function, which was originally proposed in \cite{williams1991function}, but increased its popularity with \cite{mnih2016asynchronous}, \cite{haarnoja2018soft} and \cite{schulman2017proximal}. Another option to induce exploration is adding noise perturbation, such as in \cite{burda2019exploration} or \cite{fortunato2018noisy}, adopted in well-known algorithms like \cite{hessel2018rainbow}. Recently, new approaches use intrinsic rewards and add them to the extrinsic rewards provided by the environment to encourage the agent to visit states that are still unknown or unpredictable \cite{pathak2017curiosity} \cite{badia2019never} \cite{seo2021state}. The results show that the agent is able to explore the environment even though it does not provide any extrinsic reward.

\subsection{Problem formulation} \label{sec:prob-form}

The problem is designed as a navigation task where a robot must navigate and reach the goal while avoiding collisions with a set of agents that behave like a crowd, in the shortest time possible. The simulator used is CrowdNav, widely utilized in previously mentioned works, as \cite{chen2017decentralized} \cite{everett2018motion} \cite{everett2021collision} \cite{chen2019crowd} \cite{chen2020relational} \cite{chen2020robot}  \cite{zhou2022robot}. The constraints and models of the work are defined in the way they are defined in the cited works, not defining any extra constraint to be able to make fair comparisons. The problem is defined as a crowd navigation problem with humans, but could be extensible to any kind of dynamic obstacle with a defined model, being the humans the most representative obstacles that behave as a crowd. % \cite{liu2021decentralized} \cite{hu2022crowd}.

A human's observable state consists in its position $p=[p_x,p_y]$, velocity $v=[v_x, v_y]$ and radius, $r$, and it is known by the other agents in the environment. In addition, the robot knows its own state, $w^r$, with its own position, velocity, radius, preferred velocity ($v_p$), heading angle ($\theta$) and goal coordinates ($g=[g_x, g_y]$). Having that, at time $t$, the state of the robot is $w_t^r$, the state of the human $i$ is $w_t^i$ and the state of all humans is $w_t = [w_t^1, w_t^2,...,w_t^n]$; the joint state of the environment known by the robot is $s_t=[w_t^r, w_t]$. The robot is assumed to be invisible by the humans, which means that the humans perform a reciprocal collision avoidance among themselves, but they do not try to avoid the robot.

The robot is modeled as a unicycle robot, and the set of available actions is a discrete set consisting of combinations of 16 steering angles, $\delta$, evenly spaced in $0$ and $2\pi$ radians and 5 velocities, $v$, ranging from $0$ and $v_p$. The action with null steering angle and null velocity is added to the set, resulting in the final set of 81 actions. We assume that any velocity may be achieved at any instant of time, as in previous works, to have fair comparisons. Thus, an action is defined as $a_t=[v_t, \delta_t]$ in time $t$. Finally, for each of the agents, the transition model results as follows:
\begin{equation}
\begin{split}
    & \theta_{t+1} = \theta_{t}+\delta_t, \\
    & v_t = [v_t\cos\theta_t, v_t\sin\theta_t], \\
    & p_{t+1} = [p_t+v_t\Delta t], 
\end{split}
\end{equation}
\noindent where $\Delta t$ is the time interval between consecutive time steps, set to 0.25 s.

The problem is defined as a Markov Decision Process, where the goal is to estimate the optimal policy, $\pi^*$, that chooses the optimal action, $a_t$, for a state, $s_t$, in a particular time $t$: $\pi^*:s_t\xrightarrow[]{}a_t$. The optimal policy is the one that maximizes the expected return:
\begin{equation}
    \pi^*(s_t) = \argmax_a \left(Q^*(s_t,a)\right),
\end{equation}
\noindent where $Q^*$ is the optimal action-value function, recursively defined with the Bellman equation as:
\begin{equation}
\begin{split}
    Q^*(s_t,a_t) = & \sum_{s_{t+1}} (P(s_{t+1}|s_t,a_t)[R(s_t,a_t,s_{t+1})+ \\
    & \gamma\max_{a_{t+1}}Q^*(s_{t+1},a_{t+1})]),
\end{split}
\end{equation}
\noindent where $P(s_{t+1}|s_t,a_t)$ is the probability of reaching the state $s_{t+1}$ after performing the action $a_t$ in the state $s_t$, and $R(s_t,a_t,s_{t+1})$ the reward function. The extrinsic reward provided by the environment is defined as in \cite{zhou2022robot}, encouraging the agent to navigate towards the goal, avoid collisions and maintain safe distance with humans:

\begin{equation}
  r_{ex} =
    \begin{cases}
      0.25 & \text{if goal reached}\\
      -0.25 & \text{if collision}\\
      -0.2\Delta d_g + \sum_{i=0}^Nf(\mu_i) & \text{otherwise}
    \end{cases} 
\end{equation}
\begin{equation}
    f(\mu_i) =
        \begin{cases}
          \mu_i-0.2 & \mu_i < 0.2 \\
          0 & \text{otherwise}
        \end{cases}
\end{equation}
\noindent where $d_g$ is the distance between the robot and the goal and $\mu_i$ the distance between the human $i$ and the robot, and $N$ the number of humans, being 0.2 the minimum safe distance.

\section{Approach}\label{sec:approach}

\subsection{Exploration problem}

In this work, we consider deep reinforcement learning solutions of the crowd navigation task. In the works previously mentioned that used the CrowNav simulator to develop solutions and compare them among each other, they do not add any special feature to balance exploration and exploitation, only the methods inherited by the reinforcement algorithm used. \cite{chen2017decentralized}  \cite{chen2019crowd} \cite{chen2020relational} \cite{chen2020robot}  \cite{zhou2022robot} use the decaying $\epsilon$-greedy strategy, while \cite{everett2018motion} \cite{everett2021collision} use the policy-based Asynchronous Advantage Actor-Critic algorithm, A3C, which uses a continuous action space and trains a stochastic policy, managing exploration with an entropy term that promotes action diversity.

These strategies have the risk of getting trapped in a local optima, as it will be seen in the experimentation. The amount of randomness decreases as the agent trains, due to the $\epsilon$ discount or due to the update rule (in case of the entropy term), which encourages the policy to exploit the rewards already found and converge to a solution. In any kind of problem, this could typically lead to suboptimal policies, and it is partially solved by trying different random initialization of the network weights and trying different hyperparameters values. In the crowd navigation task, the problematic increases. It is a problem almost impossible to solve by using random actions, so reward shaping is always used, and usually imitation learning for initialization. Nevertheless, both options influence the behavior of the agent, leading the agent to try to either get already known positive rewards or explore from the beginning, limiting the exploration. 

One approach to solve the problem is adding randomness directly to the neural network. Noisy networks \cite{fortunato2018noisy} proved an improvement in the scores of the Atari benchmark \cite{bellemare2013arcade}. The method consists in replacing linear layers of the neural network with noisy linear layers, whose noise is automatically tuned during training. The uncertainty in the network weights makes the decision-making variable, introducing exploration. Other approach is introducing dropout layers \cite{srivastava2014dropout} to induce variability in the action selection process and increase exploration. The use of dropout makes some number of layer outputs be ignored, randomly constructing policy subnetworks \cite{sung2018dropout}, ensuring exploration. However, the use of intrinsic rewards seemed more promising, as they explicitly encourage visiting unknown states.

\subsection{Intrinsic reward}\label{sec:intrinsic}

Intrinsic rewards are used to increase the motivation of the agent to explore new states or to reduce the uncertainty to predict the consequences of the actions. In this work, the agent would benefit from having an intrinsic reward that encourages it to visit unknown or unpredictable states until they are properly explored, and then exploit (until unknown states are found again), specially beneficial near humans and the goal.

\begin{figure}[h!]
    \centering
    \includegraphics[width=0.7\textwidth]{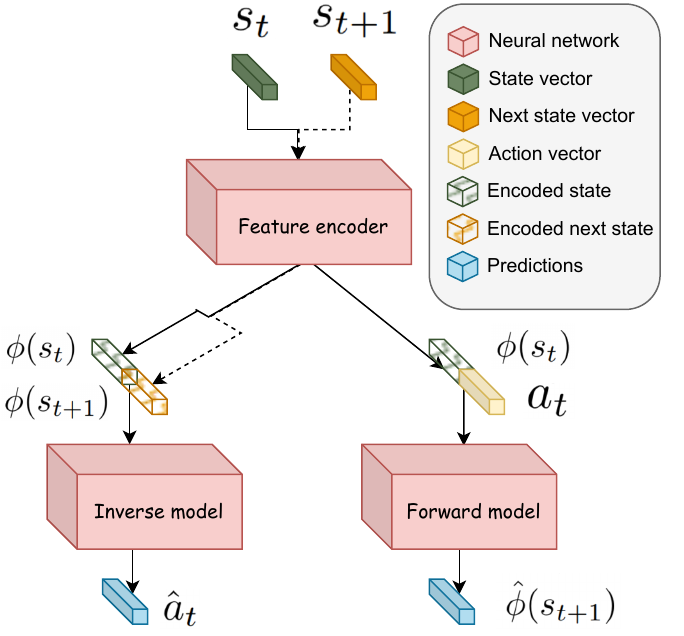}
    \caption{A diagram of the Intrinsic Curiosity Module.}
    \label{fig:curiosity}
\end{figure}

Two approaches are considered. The first one is using an Intrinsic Curiosity Module (ICM) \cite{pathak2017curiosity}, which proved to outperform other methods when it was published, and has been used as a reference in recent years. It uses a feature network, $\phi$, to encode the state, $s$, and the next state, $s_{t+1}$, to a feature space, $\phi(s)$ and $\phi(s_{t+1})$; transforming the agent-level state to a state defined by a features vector, which is the output of the network. The states in the feature spaces are used to predict the action taken, $\hat{a_t}$. The actual action taken, $a_t$, with $\phi(s)$ are used to predict the next state in the feature space, $\hat{\phi}(s_{t+1})$. This is illustrated in Figure~\ref{fig:curiosity}.

The state-of-the-art algorithms tested in the experiments are SG-D3QN \cite{zhou2022robot}, MP-RGL \cite{chen2020relational}, LSTM-RL \cite{everett2018motion} and SARL \cite{chen2019crowd}. For the feature encoder network, we include the encoder layers of the original algorithms, different for each of them (the graph attention layers for SG-D3QN, relational graph layers for MP-RGL, LSTM layers for LSTM-RL and attention layers for SARL); and two fully connected layers with 256 and 128 outputs to encode the states into vectors of size 128. The inverse model of the ICM has 3 fully connected layers with 256, 128 and 81 (number of actions) outputs and the forward model three fully connected layers with 256, 128 and 128 (encoded size) outputs. The networks are trained to minimize the prediction errors in both $\hat{a_t}$ and $\hat{\phi}(s_{t+1})$.

The intrinsic reward is computed with the Mean Squared Error (MSE) between $\phi(s_{t+1})$ and $\hat{\phi}(s_{t+1})$, being higher when the agent visits unknown or unpredictable states. The intrinsic reward, $r_{in}$, multiplied by a hyperparameter $\beta$ that controls its influence, is added to the extrinsic reward, $r_{ex}$, (the regular reward provided by the environment) to compute the total reward, $r$, and optimize the policy:

\begin{equation}
    r  = r_{ex}+\beta r_{in} = r_{ex}+\beta MSE(\phi(s_{t+1}), \hat{\phi}(s_{t+1}))
\end{equation}

The second proposed approach, the Random Encoders for Efficient Exploration (RE3) \cite{seo2021state}, maximizes the entropy of the state distribution. It uses a neural network, $f_\theta$, randomly initialized with fixed weights, $\theta$, that do not change for the whole training process of the DRL algorithm. The network represents the state in a low-dimensional space, as the previously presented feature encoder of the ICM, and has the same structure as the feature encoder of the ICM ($\phi$ of the ICM is similar to $f_\theta$). In the work, they prove that the distance between different states in the representation space is useful to quantify the similarity among the states without any training. To compute the entropy of a state, $k$-Nearest Neighbors ($k$-NN) algorithm is used, obtaining the following intrinsic reward, $r_{in}$, for a transition $i$:
\begin{equation}
    r_{in}(s_i) = \log (||f_\theta(s_i)-f_\theta(s_i)^{k-NN}||_2+1),
\end{equation}
\noindent where $f_\theta(s_i)^{k-NN}$ are the $k$-NN of $s_i$ in the representation space. The intrinsic reward is also controlled with a hyperparameter, $\beta$, as in the ICM, computing the final reward $r$ in the same way: $r = r_{ex}+\beta r_{in}$.

\section{Experiments}\label{sec:experiments}

\subsection{Environment settings} \label{sec:env-set}

The different exploration strategies are tried for several state-of-the-art algorithms, to test different models and DRL algorithms (extensible to any network or algorithm like SAC \cite{haarnoja2018soft} or PPO \cite{schulman2017proximal}). The experiments are performed in the CrowdNav environment, as explained in Section~\ref{sec:prob-form}. In each of the scenarios, 10 humans act as dynamic obstacles: 5 are set in a circle and have to cross the center of it, and 5 are randomly placed and have to cross the room (later called circle scenarios). They use the ORCA \cite{van2008reciprocal} algorithm to navigate and avoid collisions with each other. The robot is invisible for them, so that it must perform the whole collision avoidance maneuvers. When a human reaches its assigned goal, another one is randomly set, to prevent it from stopping. The agent is trained for 10000 randomly generated episodes (random position and trajectories of the agents), with the same random  weights initialization for every algorithm for a fair comparison. Videos of the experiments may be found at \url{https://github.com/dmartinezbaselga/intrinsic-rewards-navigation.git}.

\subsection{SG-D3QN}

The multistep SG-D3QN algorithm presented in \cite{zhou2022robot} is used as a referent for some of the experiments because it is a very recent work that declares to outperform the state-of-the-art. The work uses graph attention layers over a Double Dueling Deep Q-Network (D3QN) to train the agent, and $\epsilon$-greedy for exploration. It has been modified introducing noisy networks, dropout, ICM and RE3 for exploration, trying different hyperparameters and selecting the best of them in each case (a decaying dropout rate from 0.5 to 0.01 for the first 7000 training episodes, and a constant $\beta$ parameter of 0.01 for both ICM and RE3).

To validate and compare the approaches, each of the methods have been tested in 1000 random generated episodes in two different maps: Circle (Section~\ref{sec:env-set}), and Square scenarios, with 10 humans randomly placed in a squared room with trajectories that make them cross. The average returns, success rates and navigation times are gathered and presented in Table~\ref{tab:table-sg-d3qn}, as well as the same metrics obtained by a robot using ORCA \cite{van2008reciprocal} for a baseline comparison.

\begin{figure*}[]
    \centering
     \begin{tabular}[b]{c}
         \includegraphics[trim={0.5cm 0.5cm 1.75cm 1.75cm},clip,width=0.22\textwidth]{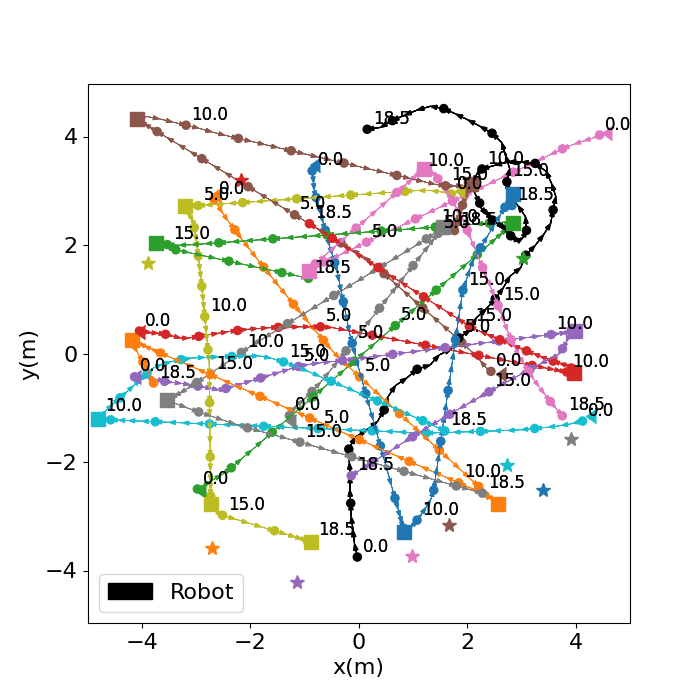}\\
         \footnotesize{(a) $\epsilon$-greedy 500 epsisodes}
    \end{tabular}
    \hfill
    \begin{tabular}[b]{c}
         \includegraphics[trim={0.5cm 0.5cm 1.75cm 1.75cm},clip,width=0.22\textwidth]{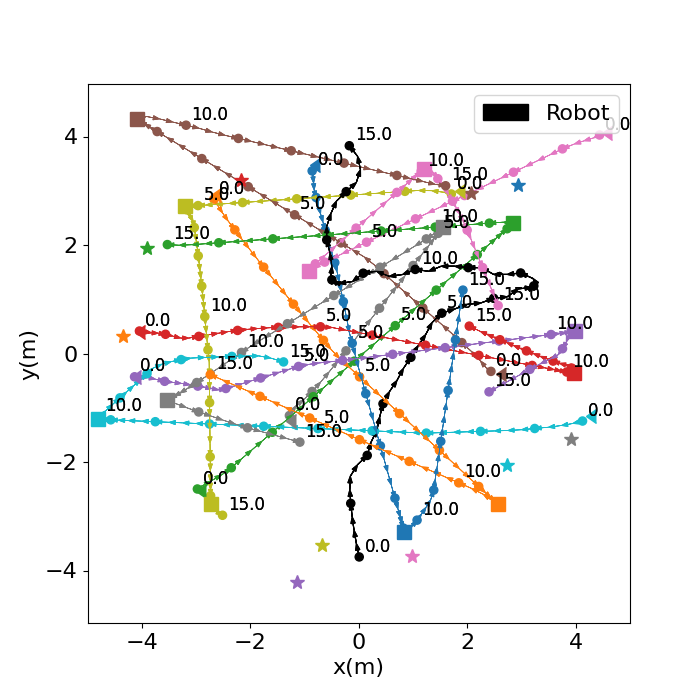} \\
         \footnotesize{(b) $\epsilon$-greedy 2500 epsisodes}
    \end{tabular}
    \hfill
    \begin{tabular}[b]{c}
         \includegraphics[trim={0.5cm 0.5cm 1.75cm 1.75cm},clip,width=0.22\textwidth]{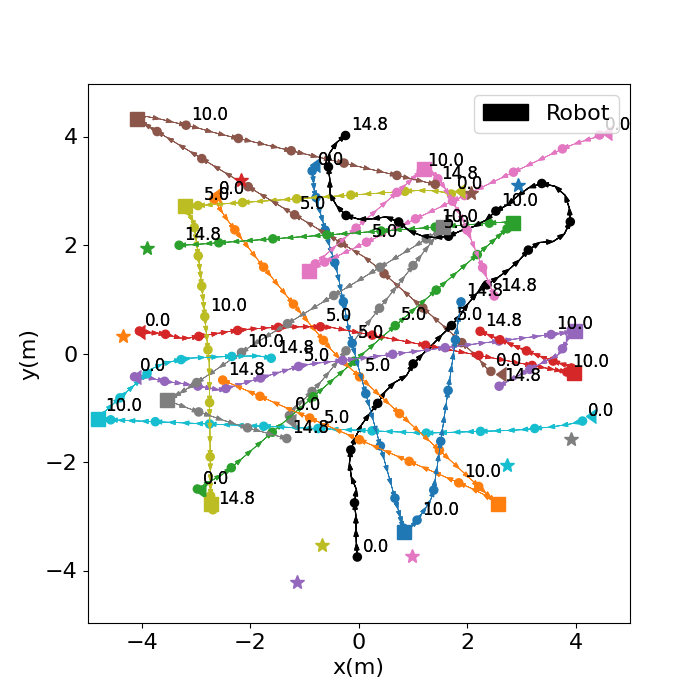} \\
         \footnotesize{(c) $\epsilon$-greedy 5000 epsisodes}
    \end{tabular}
    \hfill
    \begin{tabular}[b]{c}
         \includegraphics[trim={0.5cm 0.5cm 1.75cm 1.75cm},clip,width=0.22\textwidth]{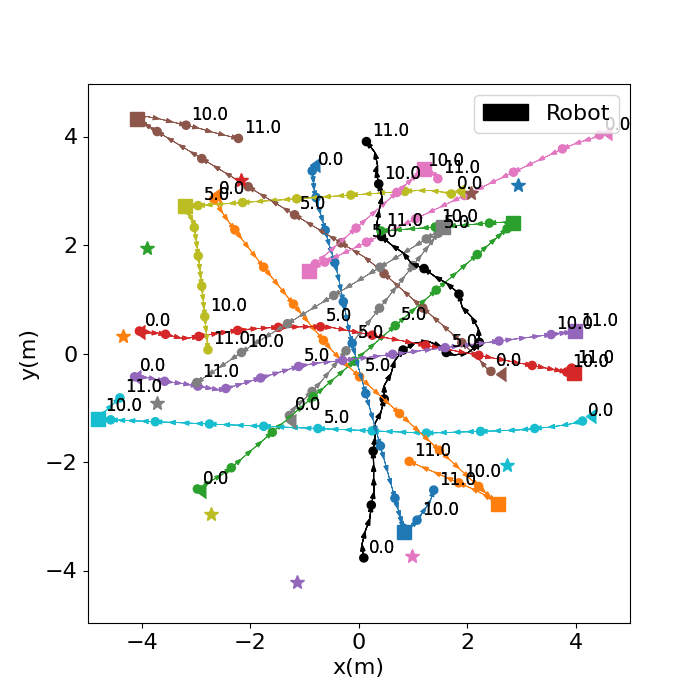}\\
         \footnotesize{(d) $\epsilon$-greedy best}
     \end{tabular}
     \hfill
     \begin{tabular}[b]{c}
        \includegraphics[trim={0.5cm 0.5cm 1.75cm 1.75cm},clip,width=0.22\textwidth]{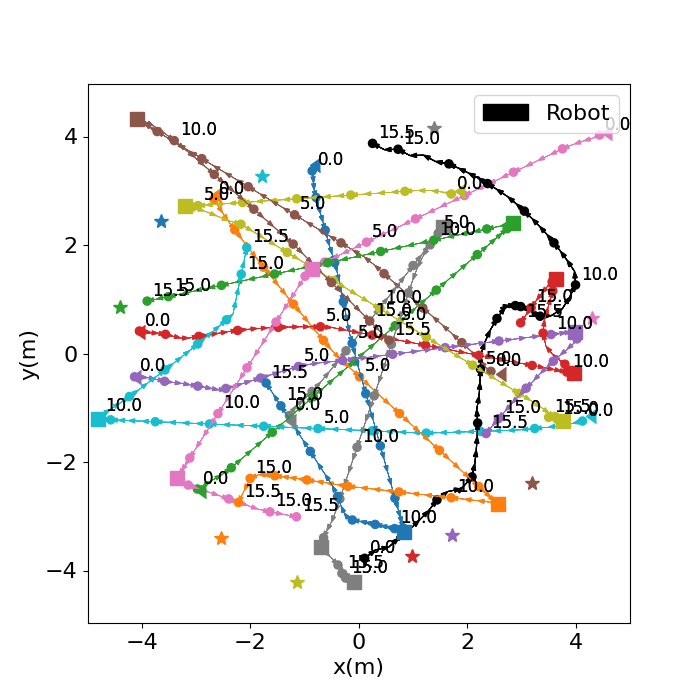}\\
         \footnotesize{(e) ICM 500 epsisodes}
    \end{tabular}
    \hfill
    \begin{tabular}[b]{c}
         \includegraphics[trim={0.5cm 0.5cm 1.75cm 1.75cm},clip,width=0.22\textwidth]{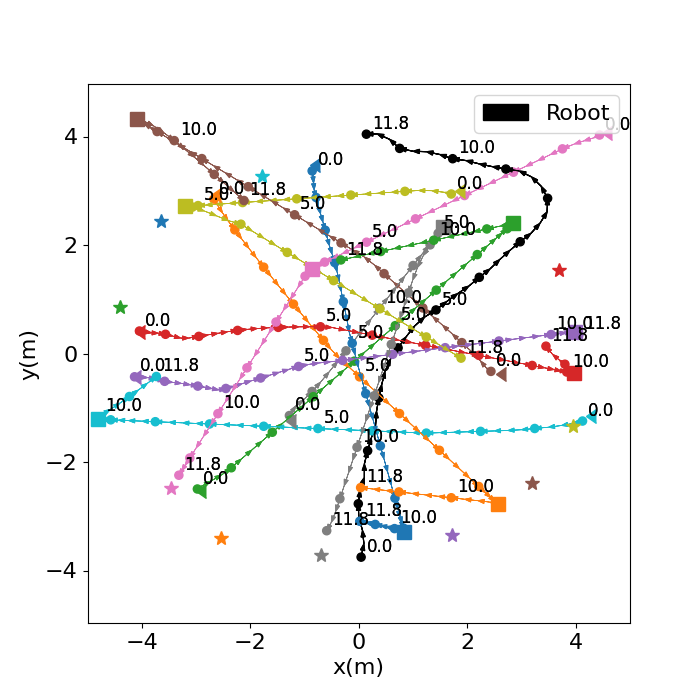}\\
         \footnotesize{(f) ICM 2500 episodes}
    \end{tabular}
    \hfill
    \begin{tabular}[b]{c}
         \includegraphics[trim={0.5cm 0.5cm 1.75cm 1.75cm},clip,width=0.22\textwidth]{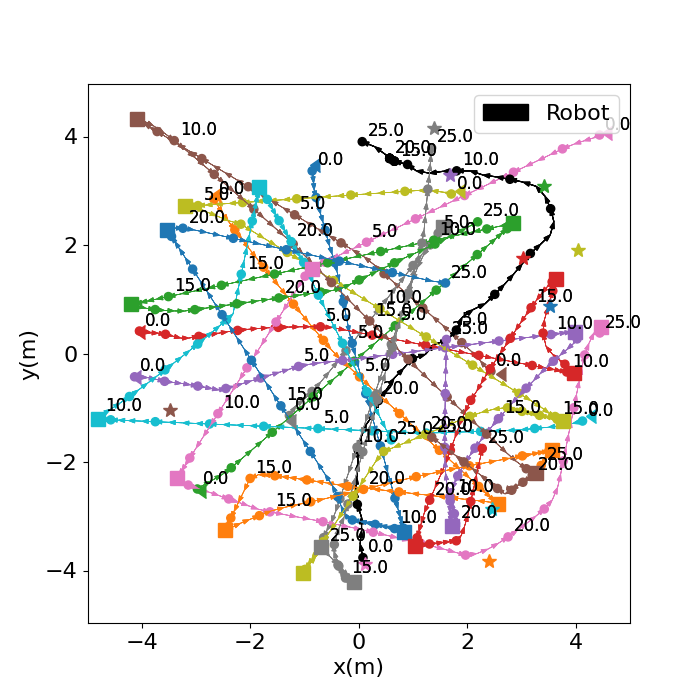}\\
         \footnotesize{(g) ICM 5000 episodes}
    \end{tabular}
    \hfill
    \begin{tabular}[b]{c}
         \includegraphics[trim={0.5cm 0.5cm 1.75cm 1.75cm},clip,width=0.22\textwidth]{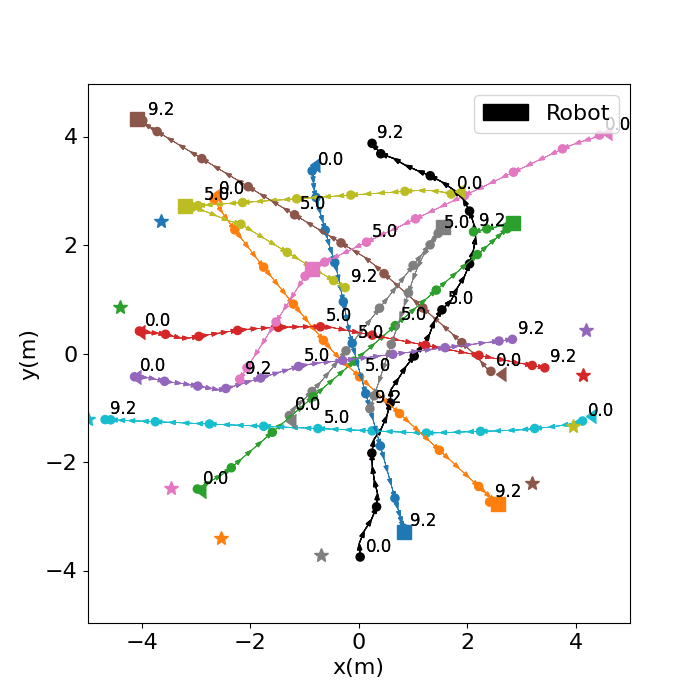} \\
         \footnotesize{(h) ICM best}
     \end{tabular}
        \caption{Comparison of robot trajectories using $\epsilon$-greedy and ICM exploration at different stages of training (500, 2500, 5000 episodes and final model).}
        \label{fig:original-vs-icm}
\end{figure*}

\begin{table}[ht]
    \centering
    %\resizebox{\textwidth}{!}{
    \begin{tabular}{|c|c|c|c|c|}
    \hline
     \multirow{2}{*}{\textbf{Scenario}} & \multirow{2}{*}{\textbf{Method}} & \textbf{Average} & \textbf{Success} & {\textbf{Navigation}}
          \\ & & \textbf{return} & \textbf{rate} & \textbf{time (sec.)}\\
         \hline
         \multirow{6}{*}{\textbf{Circle}}
          & Original & 0.6312 & 0.940 & 11.318  \\
          & Noisy & 0.5423 & 0.937 & 13.965  \\
          & Dropout & 0.6529 & 0.959 & 11.899  \\
          & ICM & 0.6680 & 0.968 & 11.366  \\ 
          & RE3 & \textbf{0.6820} & \textbf{0.971} & \textbf{11.012}  \\ \cline{2-5}
          & ORCA & 0.331 & 0.769 & 13.880 \\
          \hline
         \multirow{6}{*}{\textbf{Square}}
          & Original & 0.6396 & 0.948 & 10.545  \\
          & Noisy & 0.5566 & 0.921 & 12.742  \\
          & Dropout & 0.6484 & 0.952 & 11.148  \\
          & ICM & \textbf{0.6868} & \textbf{0.970} & 10.729  \\ 
          & RE3 & 0.6749 & 0.958 & \textbf{10.221}  \\ \cline{2-5}
          & ORCA & 0.442 & 0.840 & 12.856 \\
    \hline
    \end{tabular}
    %}

    \label{tab:table-sg-d3qn}
    \caption{Metrics of SG-D3QN (1000 random generated scenarios)}
\end{table}

The results in the table show that the intrinsic rewards clearly improve the original results and other approaches. The noisy networks have the advantage of not having any hyperparameter, but they are not enough to explore effectively the state space of the problem, as the network learns the behaviors that lead to states with safe high rewards, far from states that are near the humans, that could lead to collisions because of the noise. This prevents the robot from exploring the states that are near the humans. 

The dropout version offers very promising results despite its simplicity, but the best results are the ones obtained with intrinsic rewards, due to the fact that, they make the agent explore unknown states even though they do not seem to be promising, until they are predictable enough to estimate precisely whether they are rewarding or not, finding out behaviors and maneuvers that could not be found otherwise.

The difference in behavior is observed in the experiment shown in Figure~\ref{fig:original-vs-icm}. The trajectories performed by testing the robot in different training stages using the original $\epsilon$-greedy and the ICM strategy are plotted. The plots show the greedy trajectory of the robot and the humans until the robot reaches the goal, after training for a certain number of episodes. The scenario is the same in every plot (the humans have the same initial positions and trajectories), although the trajectories plotted are different, and that is because of the time length of the episode. The humans' intermediate goals are marked with squares and their final non-reached goals with stars. The numbers plotted near the trajectories are the simulation time at that point of the trajectory, marking them every 5 seconds and at the final time (when the robot reaches the goal). The plots show the trajectory of the robot after being trained for only 500 episodes, 2500 episodes, 5000 episodes and the final trained model (10000 episodes or the one with the best validation score); with the same SG-D3QN with the $\epsilon$-greedy strategy used for exploration or ICM.

It may be clearly seen that the $\epsilon$-greedy finds a successful trajectory from the beginning (a) and optimizes the decision-making of the agent as the training continues, resulting in a progressively refined version of the first trajectory (b), (c), (d). The trajectory obtained is increasingly  smoother and shorter in time and length in the stages shown. The robot learns to get closer to the goal and far from the humans, and, as the training continues and the exploiting part is more important, the robot exploits to get even closer to the goal until it reaches it. However, in the states where the robot is close to the goal, it will hardly explore, as exploitation is needed to reach it and, thus, the agent will be in the exploitation phase. Furthermore, the robot avoids being close to humans, as those states do not seem to be rewarding, preventing itself from learning optimal avoiding maneuvers that could imply being close to them. Therefore, it is not able to explore optimal trajectories very different from the ones already learned, and a local optima is found.

\begin{figure*}[]
    \centering
    \resizebox{0.8\textwidth}{!}{
     \begin{tabular}[b]{c}
         \includegraphics[width=0.33\textwidth]{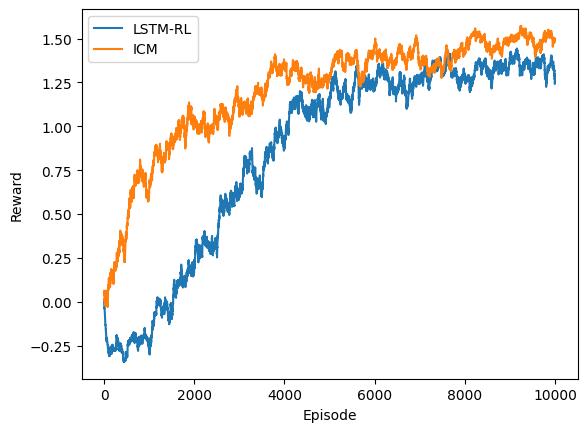}
         \includegraphics[width=0.33\textwidth]{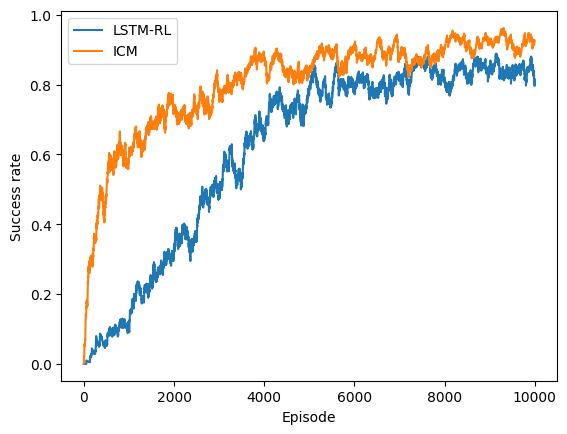}
         \includegraphics[width=0.33\textwidth]{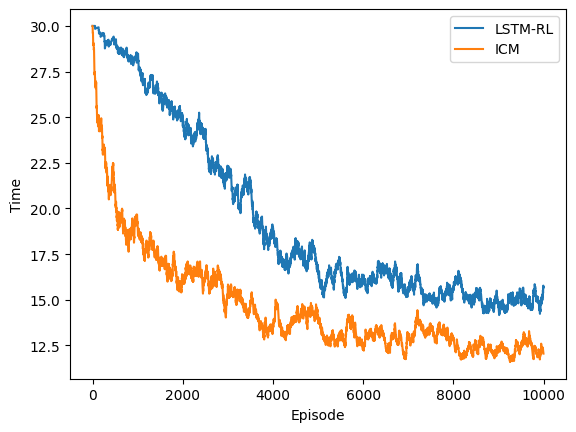}\\
         \footnotesize{(a) LSTM-RL}
    \end{tabular}
    }
    \resizebox{0.8\textwidth}{!}{
     \begin{tabular}[b]{c}
         \includegraphics[width=0.33\textwidth]{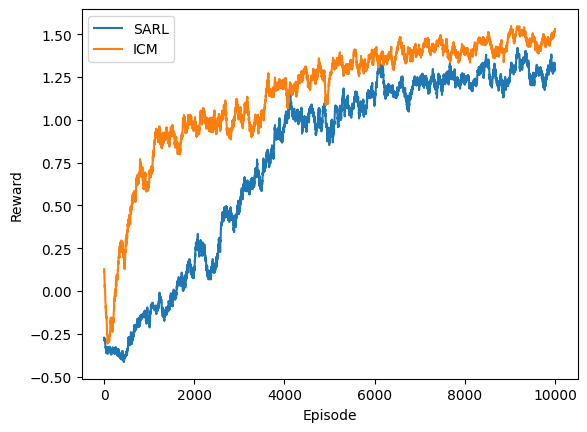}
         \includegraphics[width=0.33\textwidth]{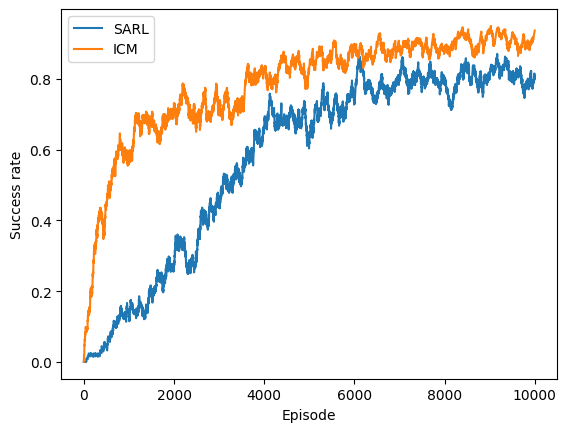}
         \includegraphics[width=0.33\textwidth]{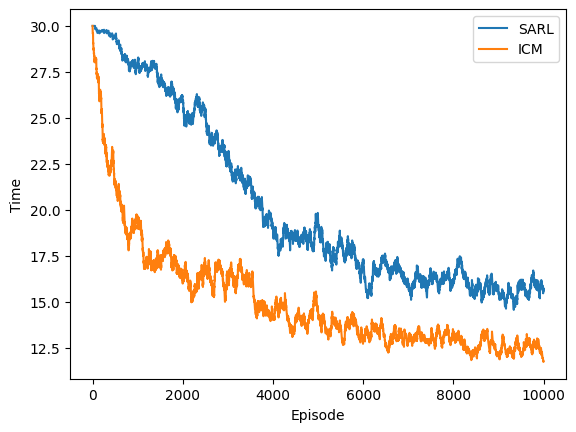}\\
         \footnotesize{(b) SARL}
    \end{tabular}
    }
    \resizebox{0.8\textwidth}{!}{
     \begin{tabular}[b]{c}
         \includegraphics[width=0.33\textwidth]{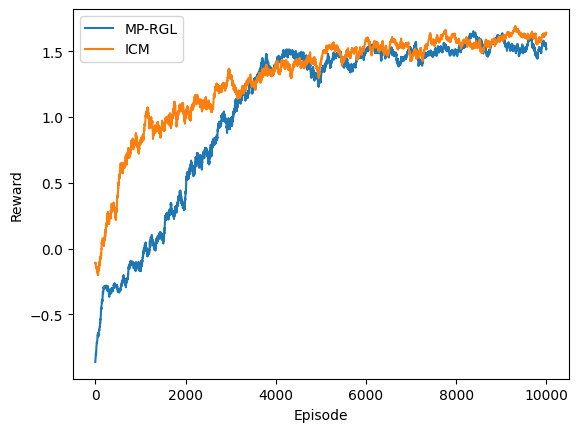}
         \includegraphics[width=0.33\textwidth]{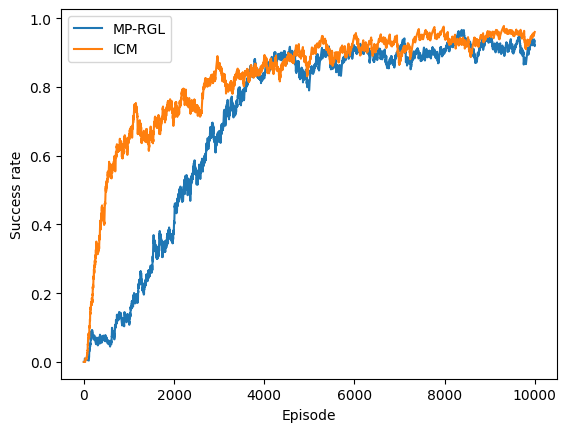}
         \includegraphics[width=0.33\textwidth]{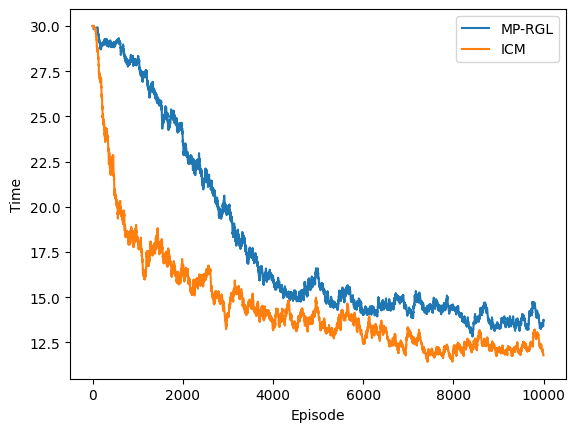}\\
         \footnotesize{(c) MP-RGL}
    \end{tabular}
    }
        \caption{Accumulated reward, success rate and navigation time metrics gathered during training, with a smoothing factor of 0.99. }
        \label{fig:lstm-sarl-mprgl}
\end{figure*}

Using the ICM intrinsic reward makes the agent to quickly find a near-optimal solution (a trajectory of only 11.8 s after 2500 training episodes (f)). Then, it keeps exploring states that may not be optimal but are uncertain and result in high intrinsic reward, resulting in trajectories that are not better than the ones of the previous stages (a trajectory of 25 s after 5000 training episodes (g), way higher than 11.8 s for 2500 training episodes (f) and 15.5 s for 500 (e)). The robot is, thus, visiting states that it has not been able to visit before, performing an effective exploration near the goal and the agents. The result is that the finally trained agent ends up predicting better the Q-values than the $\epsilon$-greedy approach, reaching the goal in only 9.2 s (h) for the 11.0 s (d) that takes the $\epsilon$-greedy version in the same fixed scenario. A similar improvement is obtained using RE3 approach, even a little bit better than the ICM in the Circle scenario.

\subsection{Other models}

The exploration strategies were tried with other well-known crowd-navigation algorithms of the state-of-the-art. Particularly, they were tried with LSTM-RL \cite{everett2018motion}, SARL \cite{chen2019crowd} and MP-RGL \cite{chen2020relational}, training the algorithms with the same setup explained in Section~\ref{sec:env-set}. Similar validation results were obtained, having the intrinsic reward strategies the best performance, and the noisy network and the originally proposed approaches the worst. 

In Figure~\ref{fig:lstm-sarl-mprgl}, the evolution of the accumulated reward (only extrinsic reward), success rate and time to reach the goal during training is shown (from left to right, respectively). The metrics of the ICM version are plotted in orange and the original one in blue. An exponential smoothing factor of 0.99 is applied to be able to interpret the curves despite their variability and take into account all past data and, thus, the speed of learning of the agent. In every graph, the metrics gathered by the original version of the algorithm and the ICM-version with a $\beta$ of 0.01 are plotted for comparison, with the smoothed metrics gathered in the 10000 training episodes. We chose the same $\beta$ as in the previous experiments because they use the same reward function (the optimal weight of $\beta$ should not differ excessively), and we wanted to prove that the approach is extensible without an extensive fine-tuning. 

For each of the three algorithms, the values of the accumulated reward and success rate of the ICM are above the ones of the original algorithm for almost every point of the graph, while for the navigation time they are clearly below. At the beginning the differences are higher, meaning that not only the intrinsic reward version has a better performance than the original one, but also the agent clearly learns faster, reaching a high reward value in about half as many episodes as the original.

\section{Conclusion}

This work presents a novel solution to improve algorithms that solve crowd-navigation problems. The work uses intrinsic rewards, which allow the agent to avoid the local optima usually found in the learning process and perform a more efficient and effective exploration. The agent tries to find states that are unknown or unpredictable until they stop being so, improving the estimation in states that are near the goal or the dynamic obstacles, resulting in shorter trajectories with higher success rates. The work has been tried in the CrowdNav simulator with other algorithms of the state-of-the-art, proving to work with numerous experiments and making the agents learn faster. The results are better that the ones obtained by other algorithms, obtaining, thus, a method that improves the existing state-of-the-art. Further work could include focusing and studying other navigation tasks to show possible improvements in related areas.

\section*{Acknowledgement}

This work was partially supported by the Spanish projects PID2019-105390RB-I00, and~Aragon Government\_FSE-T45\_20R.

%%%%%%%%%%%%%%%%%%%%%%%%%%%%%%%%%%%%%%%%%%%%%%%%%%%%%%%%%%%%%%%%%%%%%%%%%%%%%%%%
\bibliographystyle{IEEEtran}
\bibliography{IEEEabrv,references}

\begin{thebibliography}{10}
\providecommand{\url}[1]{#1}
\csname url@rmstyle\endcsname
\providecommand{\newblock}{\relax}
\providecommand{\bibinfo}[2]{#2}
\providecommand\BIBentrySTDinterwordspacing{\spaceskip=0pt\relax}
\providecommand\BIBentryALTinterwordstretchfactor{4}
\providecommand\BIBentryALTinterwordspacing{\spaceskip=\fontdimen2\font plus
\BIBentryALTinterwordstretchfactor\fontdimen3\font minus
  \fontdimen4\font\relax}
\providecommand\BIBforeignlanguage[2]{{%
\expandafter\ifx\csname l@#1\endcsname\relax
\typeout{** WARNING: IEEEtran.bst: No hyphenation pattern has been}%
\typeout{** loaded for the language `#1'. Using the pattern for}%
\typeout{** the default language instead.}%
\else
\language=\csname l@#1\endcsname
\fi
#2}}

\bibitem{hart1968formal}
P.~E. Hart, N.~J. Nilsson, and B.~Raphael, ``A formal basis for the heuristic
  determination of minimum cost paths,'' \emph{IEEE transactions on Systems
  Science and Cybernetics}, vol.~4, no.~2, pp. 100--107, 1968.

\bibitem{koenig2005fast}
S.~Koenig and M.~Likhachev, ``Fast replanning for navigation in unknown
  terrain,'' \emph{IEEE Transactions on Robotics}, vol.~21, no.~3, pp.
  354--363, 2005.

\bibitem{koenig2004lifelong}
S.~Koenig, M.~Likhachev, and D.~Furcy, ``Lifelong planning {A$\star$},''
  \emph{Artificial Intelligence}, vol. 155, no. 1-2, pp. 93--146, 2004.

\bibitem{fox1997dynamic}
D.~Fox, W.~Burgard, and S.~Thrun, ``The dynamic window approach to collision
  avoidance,'' \emph{IEEE Robotics \& Automation Magazine}, vol.~4, no.~1, pp.
  23--33, 1997.

\bibitem{fiorini1998motion}
P.~Fiorini and Z.~Shiller, ``Motion planning in dynamic environments using
  velocity obstacles,'' \emph{The International Journal of Robotics Research},
  vol.~17, no.~7, pp. 760--772, 1998.

\bibitem{lorente2018model}
M.-T. Lorente, E.~Owen, and L.~Montano, ``Model-based robocentric planning and
  navigation for dynamic environments,'' \emph{The International Journal of
  Robotics Research}, vol.~37, no.~8, pp. 867--889, 2018.

\bibitem{van2008reciprocal}
J.~Van~den Berg, M.~Lin, and D.~Manocha, ``Reciprocal velocity obstacles for
  real-time multi-agent navigation,'' in \emph{2008 IEEE International
  Conference on Robotics and Automation}.\hskip 1em plus 0.5em minus
  0.4em\relax IEEE, 2008, pp. 1928--1935.

\bibitem{berg2011reciprocal}
J.~v.~d. Berg, S.~J. Guy, M.~Lin, and D.~Manocha, ``Reciprocal n-body collision
  avoidance,'' in \emph{Robotics research}.\hskip 1em plus 0.5em minus
  0.4em\relax Springer, 2011, pp. 3--19.

\bibitem{helbing1995social}
D.~Helbing and P.~Molnar, ``Social force model for pedestrian dynamics,''
  \emph{Physical review E}, vol.~51, no.~5, p. 4282, 1995.

\bibitem{jiang2017extended}
Y.-Q. Jiang, B.-K. Chen, B.-H. Wang, W.-F. Wong, and B.-Y. Cao, ``Extended
  social force model with a dynamic navigation field for bidirectional
  pedestrian flow,'' \emph{Frontiers of Physics}, vol.~12, no.~5, pp. 1--9,
  2017.

\bibitem{trautman2010unfreezing}
P.~Trautman and A.~Krause, ``Unfreezing the robot: Navigation in dense,
  interacting crowds,'' in \emph{2010 IEEE/RSJ International Conference on
  Intelligent Robots and Systems}.\hskip 1em plus 0.5em minus 0.4em\relax IEEE,
  2010, pp. 797--803.

\bibitem{dugas2021navrep}
D.~Dugas, J.~Nieto, R.~Siegwart, and J.~J. Chung, ``Navrep: Unsupervised
  representations for reinforcement learning of robot navigation in dynamic
  human environments,'' in \emph{2021 IEEE International Conference on Robotics
  and Automation (ICRA)}.\hskip 1em plus 0.5em minus 0.4em\relax IEEE, 2021,
  pp. 7829--7835.

\bibitem{yokoyama2020autonomous}
K.~Yokoyama and K.~Morioka, ``Autonomous mobile robot with simple navigation
  system based on deep reinforcement learning and a monocular camera,'' in
  \emph{2020 IEEE/SICE International Symposium on System Integration
  (SII)}.\hskip 1em plus 0.5em minus 0.4em\relax IEEE, 2020, pp. 525--530.

\bibitem{shi2019end}
H.~Shi, L.~Shi, M.~Xu, and K.-S. Hwang, ``End-to-end navigation strategy with
  deep reinforcement learning for mobile robots,'' \emph{IEEE Transactions on
  Industrial Informatics}, vol.~16, no.~4, pp. 2393--2402, 2019.

\bibitem{tai2018socially}
L.~Tai, J.~Zhang, M.~Liu, and W.~Burgard, ``Socially compliant navigation
  through raw depth inputs with generative adversarial imitation learning,'' in
  \emph{2018 IEEE International Conference on Robotics and Automation
  (ICRA)}.\hskip 1em plus 0.5em minus 0.4em\relax IEEE, 2018, pp. 1111--1117.

\bibitem{patel2021dwa}
U.~Patel, N.~K.~S. Kumar, A.~J. Sathyamoorthy, and D.~Manocha, ``{DWA-RL}:
  Dynamically feasible deep reinforcement learning policy for robot navigation
  among mobile obstacles,'' in \emph{2021 IEEE International Conference on
  Robotics and Automation (ICRA)}.\hskip 1em plus 0.5em minus 0.4em\relax IEEE,
  2021, pp. 6057--6063.

\bibitem{chen2017decentralized}
Y.~F. Chen, M.~Liu, M.~Everett, and J.~P. How, ``Decentralized
  non-communicating multiagent collision avoidance with deep reinforcement
  learning,'' in \emph{2017 IEEE international conference on robotics and
  automation (ICRA)}.\hskip 1em plus 0.5em minus 0.4em\relax IEEE, 2017, pp.
  285--292.

\bibitem{hochreiter1997long}
S.~Hochreiter and J.~Schmidhuber, ``Long short-term memory,'' \emph{Neural
  computation}, vol.~9, no.~8, pp. 1735--1780, 1997.

\bibitem{everett2018motion}
M.~Everett, Y.~F. Chen, and J.~P. How, ``Motion planning among dynamic,
  decision-making agents with deep reinforcement learning,'' in \emph{2018
  IEEE/RSJ International Conference on Intelligent Robots and Systems
  (IROS)}.\hskip 1em plus 0.5em minus 0.4em\relax IEEE, 2018, pp. 3052--3059.

\bibitem{everett2021collision}
M.~Everett, Y.~F. Chen, and J.~P. How, ``Collision avoidance in pedestrian-rich
  environments with deep reinforcement learning,'' \emph{IEEE Access}, vol.~9,
  pp. 10\,357--10\,377, 2021.

\bibitem{chen2019crowd}
C.~Chen, Y.~Liu, S.~Kreiss, and A.~Alahi, ``Crowd-robot interaction:
  Crowd-aware robot navigation with attention-based deep reinforcement
  learning,'' in \emph{2019 International Conference on Robotics and Automation
  (ICRA)}.\hskip 1em plus 0.5em minus 0.4em\relax IEEE, 2019, pp. 6015--6022.

\bibitem{vaswani2017attention}
A.~Vaswani, N.~Shazeer, N.~Parmar, J.~Uszkoreit, L.~Jones, A.~N. Gomez,
  {\L}.~Kaiser, and I.~Polosukhin, ``Attention is all you need,''
  \emph{Advances in neural information processing systems}, vol.~30, 2017.

\bibitem{chen2020relational}
C.~Chen, S.~Hu, P.~Nikdel, G.~Mori, and M.~Savva, ``Relational graph learning
  for crowd navigation,'' in \emph{2020 IEEE/RSJ International Conference on
  Intelligent Robots and Systems (IROS)}.\hskip 1em plus 0.5em minus
  0.4em\relax IEEE, 2020, pp. 10\,007--10\,013.

\bibitem{chen2020robot}
Y.~Chen, C.~Liu, B.~E. Shi, and M.~Liu, ``Robot navigation in crowds by graph
  convolutional networks with attention learned from human gaze,'' \emph{IEEE
  Robotics and Automation Letters}, vol.~5, no.~2, pp. 2754--2761, 2020.

\bibitem{zhou2022robot}
Z.~Zhou, P.~Zhu, Z.~Zeng, J.~Xiao, H.~Lu, and Z.~Zhou, ``Robot navigation in a
  crowd by integrating deep reinforcement learning and online planning,''
  \emph{Applied Intelligence}, pp. 1--17, 2022.

\bibitem{vemula2018social}
A.~Vemula, K.~Muelling, and J.~Oh, ``Social attention: Modeling attention in
  human crowds,'' in \emph{2018 IEEE international Conference on Robotics and
  Automation (ICRA)}.\hskip 1em plus 0.5em minus 0.4em\relax IEEE, 2018, pp.
  4601--4607.

\bibitem{williams1991function}
R.~J. Williams and J.~Peng, ``Function optimization using connectionist
  reinforcement learning algorithms,'' \emph{Connection Science}, vol.~3,
  no.~3, pp. 241--268, 1991.

\bibitem{mnih2016asynchronous}
V.~Mnih, A.~P. Badia, M.~Mirza, A.~Graves, T.~Lillicrap, T.~Harley, D.~Silver,
  and K.~Kavukcuoglu, ``Asynchronous methods for deep reinforcement learning,''
  in \emph{International conference on machine learning}.\hskip 1em plus 0.5em
  minus 0.4em\relax PMLR, 2016, pp. 1928--1937.

\bibitem{haarnoja2018soft}
T.~Haarnoja, A.~Zhou, P.~Abbeel, and S.~Levine, ``Soft actor-critic: Off-policy
  maximum entropy deep reinforcement learning with a stochastic actor,'' in
  \emph{International conference on machine learning}.\hskip 1em plus 0.5em
  minus 0.4em\relax PMLR, 2018, pp. 1861--1870.

\bibitem{schulman2017proximal}
J.~Schulman, F.~Wolski, P.~Dhariwal, A.~Radford, and O.~Klimov, ``Proximal
  policy optimization algorithms,'' \emph{arXiv preprint arXiv:1707.06347},
  2017.

\bibitem{burda2019exploration}
Y.~Burda, H.~Edwards, A.~Storkey, and O.~Klimov, ``Exploration by random
  network distillation,'' in \emph{Seventh International Conference on Learning
  Representations}, 2019, pp. 1--17.

\bibitem{fortunato2018noisy}
M.~Fortunato, M.~G. Azar, B.~Piot, J.~Menick, M.~Hessel, I.~Osband, A.~Graves,
  V.~Mnih, R.~Munos, D.~Hassabis, \emph{et~al.}, ``Noisy networks for
  exploration,'' in \emph{International Conference on Learning
  Representations}, 2018.

\bibitem{hessel2018rainbow}
M.~Hessel, J.~Modayil, H.~Van~Hasselt, T.~Schaul, G.~Ostrovski, W.~Dabney,
  D.~Horgan, B.~Piot, M.~Azar, and D.~Silver, ``Rainbow: Combining improvements
  in deep reinforcement learning,'' in \emph{Thirty-second AAAI conference on
  artificial intelligence}, 2018.

\bibitem{pathak2017curiosity}
D.~Pathak, P.~Agrawal, A.~A. Efros, and T.~Darrell, ``Curiosity-driven
  exploration by self-supervised prediction,'' in \emph{International
  conference on machine learning}.\hskip 1em plus 0.5em minus 0.4em\relax PMLR,
  2017, pp. 2778--2787.

\bibitem{badia2019never}
A.~P. Badia, P.~Sprechmann, A.~Vitvitskyi, D.~Guo, B.~Piot, S.~Kapturowski,
  O.~Tieleman, M.~Arjovsky, A.~Pritzel, A.~Bolt, \emph{et~al.}, ``Never give
  up: Learning directed exploration strategies,'' in \emph{International
  Conference on Learning Representations}, 2019.

\bibitem{seo2021state}
Y.~Seo, L.~Chen, J.~Shin, H.~Lee, P.~Abbeel, and K.~Lee, ``State entropy
  maximization with random encoders for efficient exploration,'' in
  \emph{International Conference on Machine Learning}.\hskip 1em plus 0.5em
  minus 0.4em\relax PMLR, 2021, pp. 9443--9454.

\bibitem{bellemare2013arcade}
M.~G. Bellemare, Y.~Naddaf, J.~Veness, and M.~Bowling, ``The arcade learning
  environment: An evaluation platform for general agents,'' \emph{Journal of
  Artificial Intelligence Research}, vol.~47, pp. 253--279, 2013.

\bibitem{srivastava2014dropout}
N.~Srivastava, G.~Hinton, A.~Krizhevsky, I.~Sutskever, and R.~Salakhutdinov,
  ``Dropout: a simple way to prevent neural networks from overfitting,''
  \emph{The journal of machine learning research}, vol.~15, no.~1, pp.
  1929--1958, 2014.

\bibitem{sung2018dropout}
T.~T. Sung, D.~Kim, S.~J. Park, and C.-B. Sohn, ``Dropout acts as auxiliary
  exploration,'' \emph{International Journal of Applied Engineering Research},
  vol.~13, no.~10, pp. 7977--7982, 2018.

\end{thebibliography}

\end{document}